\def\paperID{11363}
\def\confName{CVPR}
\def\confYear{2024}

\def\authorBlock{
    Yichen Bai $^{1}$\thanks{Equal contribution.} \qquad
    Zongbo Han $^{1}$\footnotemark[1] \qquad
    Changqing Zhang $^{1}$\ \qquad 
    Bing Cao $^{1}$\qquad \\
    Xiaoheng Jiang $^{2}$\qquad
    Qinghua Hu $^{1}$\qquad \\
    $^{1}$Tianjin University \qquad 
    $^{2}$Zhengzhou University \\
    {\tt\small \{ycfate, zongbo, zhangchangqing, caobing, huqinghua\}@tju.edu.cn} \\
    {\tt\small \{jiangxiaoheng\}@zzu.edu.cn}
}

\newif\ifreview 
\newif\ifarxiv \newcommand{\arxiv}{\arxivtrue}
\newif\ifcamera 
\newif\ifrebuttal 

\arxiv 
\newcommand{\idlike}{\texttt{ID-like}\xspace}

\pdfoutput=1
\documentclass[10pt,twocolumn,letterpaper]{article}
\ifreview \usepackage[review]{cvpr} \fi
\ifarxiv \usepackage[pagenumbers]{cvpr} \fi
\ifrebuttal \usepackage[rebuttal]{cvpr} \fi
\ifcamera \usepackage{cvpr} \fi

\usepackage{graphicx}	
\usepackage{amsmath}	
\usepackage{amssymb}	
\usepackage{booktabs}
\usepackage{times}
\usepackage{microtype}
\usepackage{epsfig}
\usepackage[table,xcdraw,dvipsnames]{xcolor}
\usepackage{caption}
\usepackage{float}
\usepackage{placeins}
\usepackage{color, colortbl}
\usepackage{stfloats}
\usepackage{enumitem}
\usepackage{tabularx}
\usepackage{xstring}
\usepackage{multirow}
\usepackage{xspace}
\usepackage{url}
\usepackage{subcaption}
\usepackage{xcolor}
\usepackage[hang,flushmargin]{footmisc}
\usepackage{tikz}

\ifcamera \usepackage[accsupp]{axessibility} \fi

% ...

\ifarxiv  \fi

\newcommand{\R}[1]{{%
    \textbf{%
        \ifstrequal{#1}{1}{\textcolor{red}{R#1}}{%
        \ifstrequal{#1}{2}{\textcolor{blue}{R#1}}{%
        \ifstrequal{#1}{3}{\textcolor{magenta}{R#1}}{%
        \ifstrequal{#1}{4}{\textcolor{teal}{R#1}}{%
                           \textcolor{cyan}{R#1}%
        }}}}%
    }%
}}

\usepackage{xr-hyper}

\makeatletter
\newcommand*{\addFileDependency}[1]{
  \typeout{(#1)}
  \@addtofilelist{#1}
  \IfFileExists{#1}{}{\typeout{No file #1.}}
}

\makeatother

\definecolor{cvprblue}{rgb}{0.21,0.49,0.74}
\usepackage[pagebackref,breaklinks,colorlinks,citecolor=cvprblue]{hyperref}
\usepackage[capitalize]{cleveref}

\usepackage{wrapfig}
\usepackage{svg}
\usepackage{algorithm}
\usepackage{algpseudocode}
\usepackage[thicklines]{cancel}
\usepackage{multirow}
\usepackage{adjustbox}
\usepackage{tabularx}
\usepackage{subcaption}
\usepackage{amsthm}
\usepackage{booktabs}

\crefname{section}{Sec.}{Secs.}
\crefname{table}{Table}{Tables}
\crefname{figure}{Fig.}{Figs.}

\begin{document}
\title{\idlike Prompt Learning for Few-Shot Out-of-Distribution Detection}
\author{\authorBlock}
\maketitle

\begin{abstract}
Out-of-distribution (OOD) detection methods often exploit auxiliary outliers to train model identifying OOD samples, especially discovering challenging outliers from auxiliary outliers dataset to improve OOD detection. However, they may still face limitations in effectively distinguishing between the most challenging OOD samples that are much like in-distribution (ID) data, i.e., \idlike samples. To this end, we propose a novel OOD detection framework that discovers \idlike outliers using CLIP \cite{DBLP:conf/icml/RadfordKHRGASAM21} from the vicinity space of the ID samples, thus helping to identify these most challenging OOD samples. Then a prompt learning framework is proposed that utilizes the identified \idlike outliers to further leverage the capabilities of CLIP for OOD detection. Benefiting from the powerful CLIP, we only need a small number of ID samples to learn the prompts of the model without exposing other auxiliary outlier datasets. By focusing on the most challenging \idlike OOD samples and elegantly exploiting the capabilities of CLIP, our method achieves superior few-shot learning performance on various real-world image datasets (e.g., in 4-shot OOD detection on the ImageNet-1k dataset, our method reduces the average FPR95 by 12.16\% and improves the average AUROC by 2.76\%, compared to state-of-the-art methods). 
\end{abstract}
\section{Introduction}
\label{sec:intro}


\ 

When deploying machine learning models in practical settings, it is possible to come across out-of-distribution (OOD) samples that were not encountered during training. The risk of incorrect decisions rises when it comes to these OOD inputs, which could pose serious safety issues, particularly in applications like autonomous driving and medical diagnosis. The system needs to identify OOD samples in addition to performing well on ID samples in order to produce trustworthy predictions. OOD detection is therefore quite critical for safely deploying machine learning models in reality.


\setlength{\belowcaptionskip}{-12pt}
\begin{figure}[tp]
  \centering
  \includegraphics[width=0.45\textwidth]{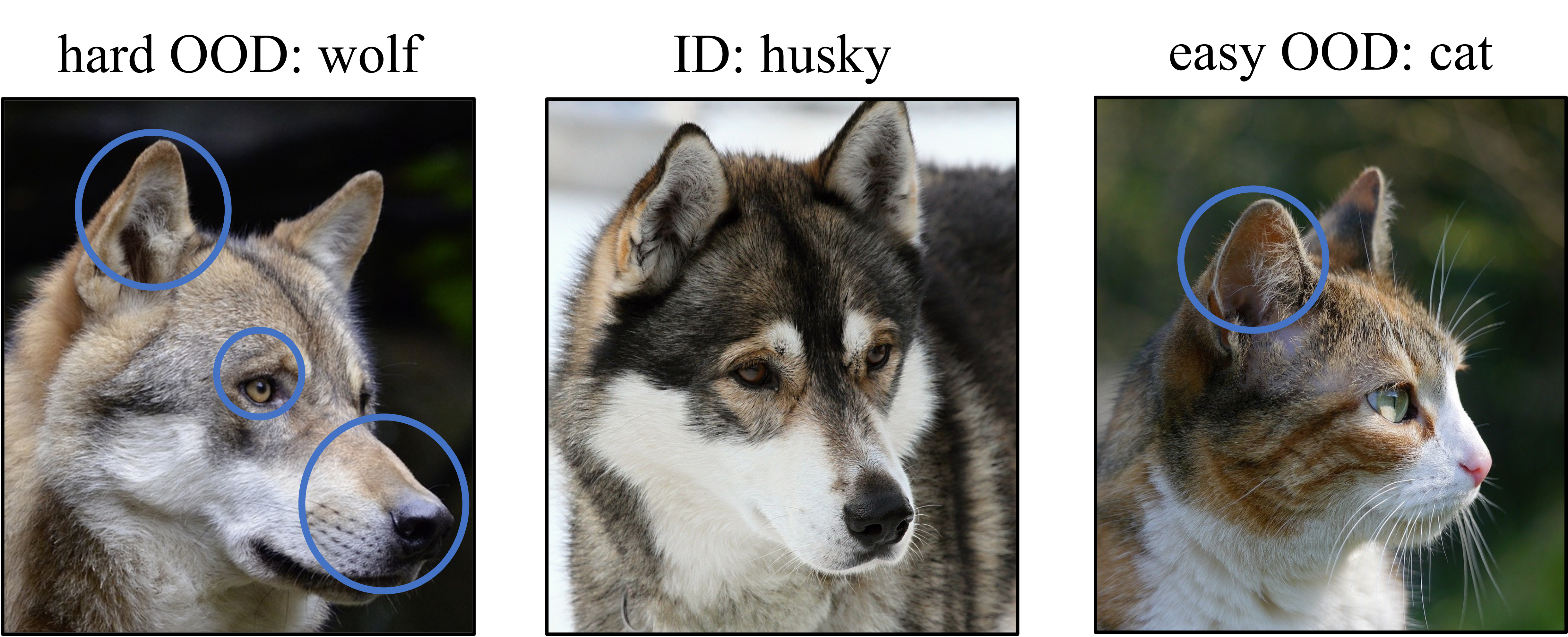}
  \caption{
    Hard and easy OOD examples: hard OOD samples typically contain more features correlated to ID samples, i.e., they behave more \idlike.
    }
    \label{fig_challenging_sample}
    \vspace{-0.15em}
\end{figure}

Existing methods \citep{DBLP:conf/iclr/HendrycksG17, DBLP:conf/nips/LeeLLS18, DBLP:conf/iclr/LiangLS18} usually focus on detecting OOD examples only using ID data in training to predict lower confidence \citep{DBLP:conf/cvpr/0001AB19, DBLP:conf/iclr/Meinke020} or higher energy \citep{DBLP:conf/nips/LiuWOL20} for OOD samples. However, due to the lack of OOD information, these models struggle to be effective in OOD detection. Therefore, some studies \citep{DBLP:conf/nips/LiuWOL20, DBLP:conf/iclr/HendrycksMD19} suggest using auxiliary outliers to regularize the models and identity OOD samples. \citet{DBLP:conf/pkdd/ChenLWLJ21} and \citet{DBLP:conf/icml/MingFL22} suggested that selecting more challenging outlier samples can help the model learn a better decision boundary between ID and OOD. However, these limited auxiliary outliers contain even less challenging outliers. Furthermore, most of these methods require additional outlier data, which makes them ineffective when outlier datasets are unavailable. Recently, \citet{DBLP:conf/iclr/DuWCL22} proposed to synthesize virtual outlier data from the low-likelihood region in the feature space of ID data to construct outliers during training without additional data. This method shows strong efficacy in distinguishing between ID and OOD. However, there are two main limitations: i) it assumes that ID data in the feature space conforms to a class conditional Gaussian distribution, which does not always hold in the complex real-world applications \citep{DBLP:conf/iclr/TaoDZ023}; 
ii) it requires numerous ID samples to construct a more accurate distribution of ID data, while obtaining a large number of ID samples is often costly. 
Accordingly, in this work, we focus on flexibly constructing informative outliers to improve the identification of challenging OOD samples.

\setlength{\abovecaptionskip}{-2pt}
\setlength{\belowcaptionskip}{-8pt}
\begin{figure*}[tp]
    \begin{center}
    \includegraphics[width=\textwidth]{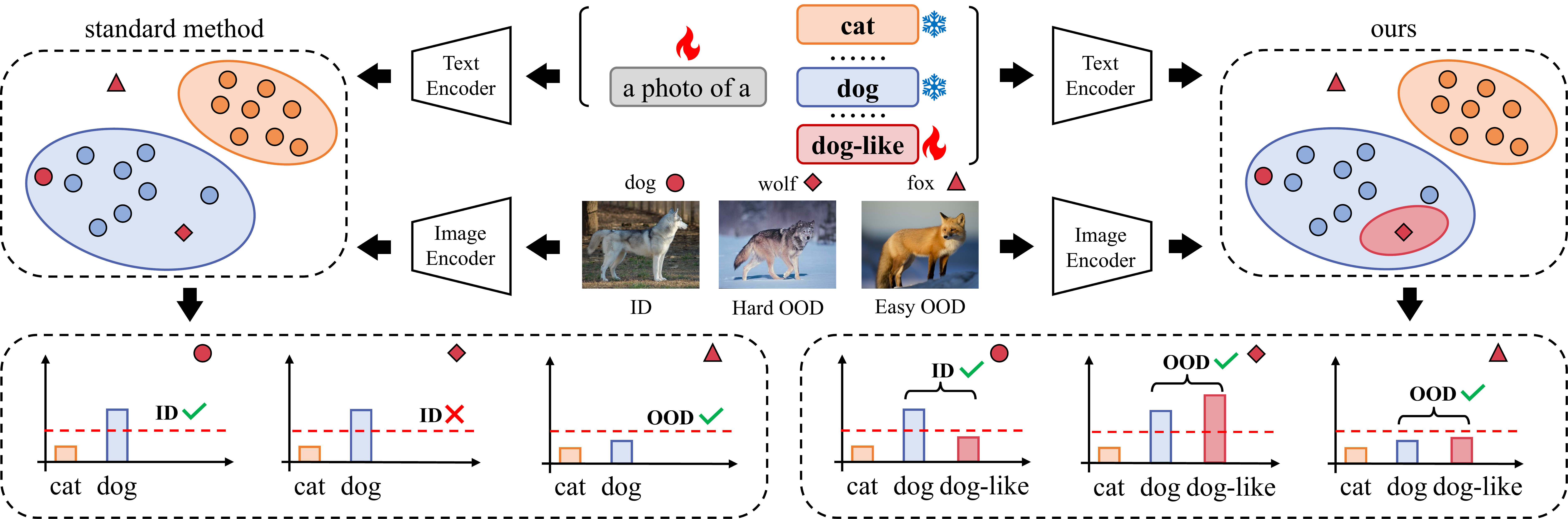}
    \end{center}
    \caption{
    The standard method can only output the predicted probabilities of samples for each ID class. In contrast, our approach can automatically learn additional classes that are highly correlated but distinct from the ID classes, thereby effectively identifying challenging \idlike OOD samples. (Note that the dog-like prompt in the figure is learnable.)
    }
    \label{fig_motive}
\end{figure*}

In this paper, we first construct outliers highly correlated with ID data and introduce a novel \idlike prompts for OOD detection, thereby effectively identifying challenging OOD samples. We find that challenging OOD samples often behave highly correlated with ID data, exhibiting high visual or semantic similarity, e.g., the local feature of OOD being relevant to ID (as shown in Fig.~\ref{fig_challenging_sample}). Since these \idlike features of OOD samples lead to erroneous predictions, a natural idea arises: extracting relevant features from ID samples to construct challenging OOD samples. 
To this end, we perform multiple samplings on vicinity space of ID samples. Among these samplings, those with lower similarity to the ID prompts are not classified as ID classes, even they contain the features correlated with ID classes. Therefore, these samples are naturally selected as challenging OOD samples. Differing from VOS \citep{DBLP:conf/iclr/DuWCL22} and NPOS \citep{DBLP:conf/iclr/TaoDZ023}, which synthesize virtual outliers in low-likelihood regions of the feature space, our method constructs outliers directly from the original images, enhancing the flexibility and interpretability. 


Although we can construct challenging OOD samples, it is still challenging to effectively identify these OOD samples. As shown in the left part of Fig.~\ref{fig_motive}, ``\textit{wolf}'' represents a challenging OOD example of ``\textit{dog}'' class. These images are similar to ID prompts, resulting in high classification probabilities and significant challenges in distinguishing between ID and OOD. We argue that relying solely on ID prompts is insufficient to address this issue. Therefore, we introduce additional prompts to enhance OOD identification. As shown in the right part of Fig.~\ref{fig_motive}, we develop an additional prompt, termed ``\textit{dog-like}'', which is similar to the prompt of ``\textit{dog}''. If we can increase the similarity between the ``\textit{dog-like}'' prompt and OOD samples that are highly correlated with ``\textit{dog}'', the model would recognize dogs through the ``\textit{dog}'' prompt and identify challenging OOD samples (including ``\textit{wolf}'') through the ``\textit{dog-like}'' prompt. Specifically, we align the additional prompts with these constructed challenging OOD, creating OOD prompts similar to ID prompts to effectively identify challenging OOD samples.
Extensive experiments demonstrate that our method achieves superior few-shot OOD detection performance on a wide variety of real-world tasks. 
Compared to methods~\citep{DBLP:conf/iclr/TaoDZ023, DBLP:conf/iclr/DuWCL22} that require a large amount of data during training, our method significantly reduces the average FPR95 score from 38.24\% to 24.08\% and improves the average AUROC from 91.60\% to 94.70\% even using only one image for each class. 
We summarize our main contributions as follows: 
\begin{itemize}
    \item 
    We propose a novel framework without additional training to automatically explore \idlike OOD samples in the vicinity space of ID samples by leveraging CLIP, which assists the model in effectively identifying challenging OOD samples correlated to the ID.
    \item 
     By exploiting the capacity of a pre-trained visual-language model, an \idlike prompt learning method is proposed to identify the most challenging OOD samples, which behave \idlike yet are distinct. 
    \item 
    We validated our method on several large-scale datasets, and the results show that our method achieved impressive performance, with an average AUROC of 96.66\% in 4-shot OOD detection on ImageNet-1K. Additional ablation experiments are also conducted to demonstrate the effectiveness of the designed approach.
\end{itemize}

\section{Related Work}
\label{sec:related}

\ 

\noindent\textbf{OOD Detection with Pre-trained Vision-language Models. }
\citet{DBLP:conf/iclr/HendrycksG17} established a baseline for OOD detection using the maximum softmax probability (MSP). Subsequent works have explored OOD detection via ODIN scores \citep{DBLP:conf/iclr/LiangLS18, DBLP:conf/cvpr/HsuSJK20} and Mahalanobis scores \citep{DBLP:conf/nips/LeeLLS18}. \citet{DBLP:conf/nips/FortRL21} first extended the OOD detection task to pre-trained vision-language models. 
\citet{DBLP:conf/aaai/Esmaeilpour00022} enhanced the OOD detection performance of pre-trained vision-language models by generating additional negative labels to construct negative prompts. Recently, \citet{DBLP:conf/nips/MingCGSL022} extended MSP to pre-trained vision-language models and explored the impact of softmax and temperature scaling on OOD detection. CLIPN \citep{Wang_2023_ICCV} fine-tuned CLIP to enable it to output negative prompts to assess the probability of a concept not being present in the image. 

\noindent\textbf{Contrastive Vision-language Models. }
Compared to traditional machine learning models, recent large-scale pre-trained vision-language models have achieved great progress in various downstream tasks. For instance, CLIP \citep{DBLP:conf/icml/RadfordKHRGASAM21}, FILIP \citep{ DBLP:conf/iclr/YaoHHLNXLLJX22} and ALIGN \citep{DBLP:conf/icml/JiaYXCPPLSLD21} leverages contrastive loss, such as InfoNCE loss \citep{DBLP:journals/corr/abs-1807-03748}, to learn aligned representations of images and text. The representation distance of matching image-text pairs becomes closer while those of non-matching pairs are farther apart. Specifically, these methods employ a straightforward dual-stream architecture comprising an image encoder and a text encoder, which maps image and text features into a shared space for similarity computation. CLIP \citep{DBLP:conf/icml/RadfordKHRGASAM21} benefits from a curated dataset of 400 million image-text pairs, and ALIGN \citep{DBLP:conf/icml/JiaYXCPPLSLD21} utilizes 1.8 billion pairs of noisy image-text data. Due to the large-scale paired data, these vision-language pre-trained models demonstrate impressive performance. 

\noindent\textbf{CLIP-based Prompt Learning.}
In Natural Language Processing (NLP), \citet{DBLP:conf/emnlp/PetroniRRLBWM19} conceptualized prompting as akin to a fill-in-the-blanks task. 
The core idea is to induce a pre-trained language model to generate answers given cloze-style prompts. Tasks such as sentiment analysis can be effectively addressed using this paradigm. However, it relies heavily on a well-designed prompt. To avoid manually designing a large number of prompts, some studies \citep{DBLP:conf/acl/LiL20, DBLP:conf/emnlp/LesterAC21} introduce prompt tuning as a solution. Prompt tuning learns the prompt from downstream data in the continual input embedding space, which presents a parameter-efficient way of fine-tuning foundation models. Despite the widespread adoption of prompt learning within NLP, its exploration within the visual domain remains limited. Recently, CoOp \citep{DBLP:journals/ijcv/ZhouYLL22} and CoCoOp \citep{DBLP:conf/cvpr/ZhouYL022} apply prompt tuning to CLIP \citep{DBLP:conf/icml/RadfordKHRGASAM21}, which tune prompts via minimizing the classification loss on the target task and effectively improves CLIP's performance on the corresponding downstream tasks. 
Plenty of studies \citep{DBLP:conf/cvpr/LuLZL022, DBLP:conf/nips/ShuNHYGAX22, DBLP:journals/corr/abs-2306-01293} leverage prompt learning based on CLIP to enhance performance across various downstream tasks.

\section{Method}
\label{sec:method}


\begin{figure*}[tp]
    \begin{center}
    \includegraphics[width=0.9\textwidth]{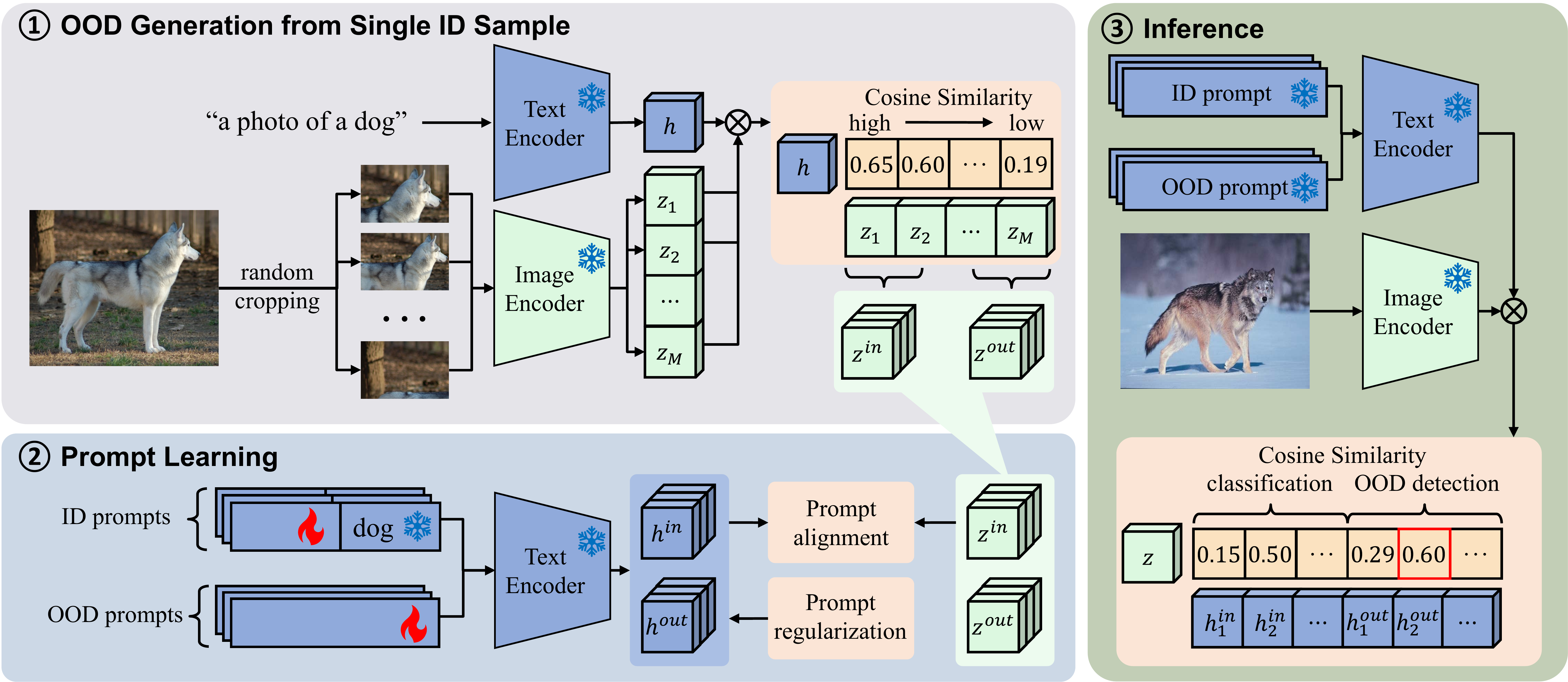}
    \end{center}
    \caption{
    Overview of our method. We conduct multiple random cropping on ID sample and filter them based on their cosine similarity with established ID zero-shot prompts, thereby generating both ID and OOD data. Subsequently, prompt learning is employed to acquire prompts corresponding to the ID and \idlike OOD samples. The obtained prompts can effectively identify OOD samples in the inference stage.
    }
    \label{fig_framework}
\end{figure*}

\subsection{Preliminaries}

\textbf{Zero-shot classification with CLIP. }
CLIP consists of a text encoder $\displaystyle \mathcal{T} : t \to \mathbb{R}^{d}$ and an image encoder  $\displaystyle \mathcal{I} : x \to \mathbb{R}^{d} $, which are used to obtain the feature vectors of text $t$ and image $x$, respectively. When performing a classification task, assuming the known label set $\displaystyle \mathcal{Y}=\{y_1, y_2,...,y_K\}$, we can construct a collection of concept vectors $\displaystyle \mathcal{T}(t_k), k \in \{1, 2, ..., K\},$ where $\displaystyle t_k$ is the text prompt ``a photo of a 〈$\displaystyle y_k$〉'' for a label $\displaystyle y_k$. 
We denote the features of text and images as $\displaystyle h=\mathcal{T}(t)$ and $\displaystyle  z=\mathcal{I}(x)$, respectively. 
We first obtain the similarity of image features relative to all text features $\displaystyle s_k(x)=sim(h_{k}, z)=sim(\mathcal{T}(t_k), \mathcal{I}(x))$, where $\displaystyle sim(\cdot,\cdot)$ denotes the cosine similarity. The predicted probability $p_k$ corresponding to $y_k$ on $\displaystyle x $ can be expressed as
\begin{equation}
    \displaystyle p_k(x; \mathcal{Y}, \mathcal{T}, \mathcal{I})=
    \frac {e^{s_k(x)/\tau}} 
    {\Sigma^K_{k=1}e^{s_k(x)/\tau}},  
\end{equation}
where $\displaystyle \tau $ is the temperature of the softmax function.

\noindent\textbf{Prompt Learning.}
To further improve the performance of CLIP on few-shot classification, CoOp \citep{DBLP:journals/ijcv/ZhouYLL22} constructs a learnable tensor on the embedding layer of the text. Specifically, CoOp initializes the learnable tensor of prompt as $\displaystyle t=[V]_1[V]_2...[V]_L[CLASS]$, where $\displaystyle [V]_l (l\in\{1, 2, ..., L\})$ is a learnable vector with the same dimension as the word embedding, and the dimension is set to 512. 
Then a loss function e.g., cross-entropy loss, can be constructed to optimize the learnable prompt according to classification probability of few-shot examples.

\noindent\textbf{Out-of-Distribution Detection. }
The OOD detection usually constructs an OOD detector denoted as $\displaystyle F(x) $, i.e., a binary classifier
\begin{equation}
    \displaystyle 
    F(x)=
    \left\{
    \begin{aligned}
    ID,&    &S(x)\geq\gamma\\
    OOD,&    &S(x)<\gamma,\\
    \end{aligned}
    \right.
    \label{OOD_func}
\end{equation}
where $\displaystyle S(x) $ is a score function in OOD detection task, and $\displaystyle \gamma $ is a threshold to decide whether the samples belong to ID or OOD. For example, \citet{DBLP:conf/iclr/HendrycksG17} and \citet{DBLP:conf/nips/LiuWOL20} use the maximum classification probability of softmax and energy as the score function  $\displaystyle S(x) $, respectively.

\subsection{ID-like Prompt Learning }

\ 

In this paper, we introduce a novel model for few-shot OOD detection, which employs cropping and the CLIP model to create challenging outliers to improve the OOD detection ability. Additionally, we employ prompt learning to acquire \idlike OOD prompts. 
As shown in Fig.~\ref{fig_framework}, our framework consists of two main components:
\textbf{(1) Constructing outliers from ID samples:} The training set under the few-shot setting is represented as $\displaystyle \mathcal{D}=\{(x_1, y_1), (x_2, y_2),...,(x_N, y_N)\}$. To suffiiently explore vicinal space of training samples, we perform multiple random cropping on each ID sample $\displaystyle x_i $ to obtain the set $\displaystyle X^{crop}_i=\{x^{crop}_{i,1}, x^{crop}_{i,2}, ..., x^{crop}_{i,M}\} $, where $\displaystyle M$ is the number of random cropping iterations. Concurrently, we create corresponding class description text $\displaystyle t_k $ using predefined templates, such as ``a photo of a 〈$\displaystyle y_k$〉'', where $\displaystyle y_k \in \mathcal{Y}$ represents the corresponding class name. Subsequently, leveraging the pre-trained CLIP model, we calculate the cosine similarity between the samples in set $\displaystyle X^{crop}_i $ and the descriptions $\displaystyle t_k$. Based on the strength of cosine similarity, we then respectively extract ID and OOD samples from the highest and lowest similarity segments, defining them as $\displaystyle X^{in}_i=\{x^{in}_{i,1}, x^{in}_{i,2}, ..., x^{in}_{i,Q}\} $ and $\displaystyle X^{out}_i=\{x^{out}_{i,1}, x^{out}_{i,2}, ..., x^{out}_{i,Q}\} $, where $\displaystyle Q$ is a user-defined hyperparameter. In the end, we obtain the $\displaystyle D^{in}=\{(x^{in}_{1,1}, y_1), (x^{in}_{1,2}, y_1),...,(x^{in}_{N,Q}, y_N)\}$ and $\displaystyle D^{out}=\{x^{out}_{1,1}, x^{out}_{1,2},...,x^{out}_{N,Q}\}$, constructed from all the ID samples. 
\textbf{(2) Prompt learning:} 
We initialize a learnable prompt for each class, forming the ID prompts set $\displaystyle T^{in}=\{t^{in}_1, t^{in}_2,...,t^{in}_K\} $, and initialize an additional set of OOD prompts, $\displaystyle T^{out}=\{t^{out}_1, t^{out}_2,...,t^{out}_C\} $, where $\displaystyle C $ is the number of OOD prompts. Given the limited scope covered by individual descriptions, we introduce multiple OOD descriptions to enhance the coverage. Using CoOp \citep{DBLP:journals/ijcv/ZhouYLL22}, we initialize embeddings for these text descriptions randomly and optimize them using a loss function, thereby learning improved prompts.

\subsection{Loss Functions}

\ 

During training, we obtain in-distribution and out-of-distribution data, denoted as $\displaystyle D^{in}$ and $\displaystyle D^{out}$, based on the algorithm mentioned in the previous section. We optimize prompts through a loss function that consists of three terms. 

\noindent\textbf{In-distribution loss.} To ensure classification performance on the in-distribution data, we utilize a standard cross-entropy loss function, which measures the divergence between the predicted label probabilities and ground truth labels for ID samples.
Formally, the ID cross-entropy loss $\displaystyle \mathcal{L}_{in}$ is defined as:
\begin{equation}
    \displaystyle \mathcal{L}_{in} = 
    \mathbb{E}_{(x,y) \sim D^{in}}
    [-\log{
        \frac
        {
            e^{s_{*}/\tau}
            }
        {
            \Sigma^K_{k=1}{e^{s^{in}_k/\tau}} + 
            \Sigma^C_{c=1}{e^{s^{out}_c/\tau}}
        }
    }],
    \label{L_id}
\end{equation}
where $\displaystyle s_{*}=sim(\mathcal{T}(t_*), \mathcal{I}(x))$, $\displaystyle s^{in}_k=sim(\mathcal{T}(t^{in}_k), \mathcal{I}(x))$, $\displaystyle s^{out}_c=sim(\mathcal{T}(t^{out}_c), \mathcal{I}(x))$, $\displaystyle t_*$ represents the features of textual description of ground-truth label $\displaystyle y_*$ corresponding to $\displaystyle x$, $\displaystyle t^{in}_k \in T^{in}$ and $\displaystyle t^{out}_c \in T^{out}$. 

\noindent\textbf{Out-of-distribution loss.} To align OOD prompts with outliers, we use the cross-entropy loss function. It is important to note that in an ideal scenario, each category would have an ID prompt and an OOD prompt. However, to conserve computational resources and enhance training efficiency, we have fixed the number of OOD prompts at $100$. 
Consequently, when there are insufficient OOD prompts to establish a one-to-one correspondence with the ID categories, we maximize the holistic similarity between the OOD prompts and outliers. 
To accomplish this, we propose the following loss $\displaystyle \mathcal{L}_{out}$:
\begin{equation}
    \displaystyle \mathcal{L}_{out}=
    \mathbb{E}_{x \sim D^{out}}
    [-\log{
        \frac
        {
            \Sigma^C_{c=1}{e^{s^{out}_c/\tau}}
            }
        {
            \Sigma^K_{k=1}{e^{s^{in}_k/\tau}} + 
            \Sigma^C_{c=1}{e^{s^{out}_c/\tau}}
        }
    }].
    \label{L_out}
\end{equation}

Additionally, we observed that implementing $\displaystyle \mathcal{L}_{out}$ during training in the following form is more conducive to optimizing prompts:
\begin{equation}
    \displaystyle \mathcal{L}_{out}=
    \mathbb{E}_{x \sim D^{out}}
    [\log{
        \frac
        {
            \Sigma^K_{k=1}{e^{s^{in}_k/\tau}}
            }
        {
            \Sigma^K_{k=1}{e^{s^{in}_k/\tau}} + 
            \Sigma^C_{c=1}{e^{s^{out}_c/\tau}}
        }
    }].
    \label{L_out_}
\end{equation}

Although their optimization goals are similar, the former tends to maximize the similarity between OOD prompts and outliers, while the latter tends to minimize the similarity between ID prompts and outliers, resulting in slight differences during training. 

\noindent\textbf{Diversity regularization.} Since all OOD prompts are randomly initialized and optimized under the same objective shown in Eq.~\ref{L_out} , there arises a risk of excessive similarity between OOD prompts. Similar OOD prompts may lead to a reduction in the number of detectable OOD classes. To mitigate this issue and ensure the diversity of OOD prompts, we introduce an additional loss $\mathcal{L}_{div}$ that explicitly maximizes the dissimilarity between prompts:
\begin{equation}
    \displaystyle \mathcal{L}_{div}=
    \frac
    {
        \Sigma^{C-1}_{c=1}\Sigma^{C}_{j=c+1}sim(h^{out}_{c}, h^{out}_{j})
    }
    {
        C(C-1)/2
    },
    \label{L_diff}
\end{equation}
where $ h^{out}_c=\mathcal{T}(t^{out}_c), h^{out}_j=\mathcal{T}(t^{out}_j)$. $\displaystyle t^{out}_c, t^{out}_j \in T^{out}$ denote the $c$-th and $j$-th prompt in the OOD prompts. $\displaystyle sim(\cdot,\cdot)$ denotes the cosine similarity.

The overall loss function with balanced hyperparameter $\lambda_{out}$ and $\lambda_{div}$ is:
\begin{equation}
\mathcal{L}=\mathcal{L}_{in}+\lambda_{out} \mathcal{L}_{out}+\lambda_{div} \mathcal{L}_{div}.
\end{equation}


\noindent\textbf{Inference.} When performing the classification task, we utilize the same classification method as CLIP, relying solely on the ID prompts for classification. For OOD detection, we define the scoring function as:
\begin{equation}
    \displaystyle 
    S(x)=\frac
        {
            \Sigma^K_{k=1}{e^{s^{in}_k/\tau}}
            }
        {
            \Sigma^K_{k=1}{e^{s^{in}_k/\tau}} + 
            \Sigma^C_{c=1}{e^{s^{out}_c/\tau}}
        }.
\end{equation}
\section{Experiments}
\label{sec:experiments}

\subsection{Experimental Setup}

\noindent\textbf{Datasets.}
Different from previous OOD detection tasks, we mainly aim to achieve OOD detection in the open-world setting, so we do not choose some toy (e.g., low-resolution) datasets, such as CIFAR \citep{Krizhevsky2009LearningML} and MNIST \citep{DBLP:conf/nips/Lakshminarayanan17}. In our work, we follow the settings of MOS \citep{DBLP:conf/cvpr/HuangL21} and MCM \citep{DBLP:conf/nips/MingCGSL022}, which use ImageNet-1k \citep{DBLP:conf/cvpr/DengDSLL009} as ID data and a subset of iNaturalist \citep{DBLP:conf/cvpr/HornASCSSAPB18}, PLACES \citep{DBLP:journals/pami/ZhouLKO018} and TEXTURE \citep{DBLP:conf/cvpr/CimpoiMKMV14} as OOD data. 
SUN \citep{DBLP:conf/cvpr/XiaoHEOT10} is tested independently as a specific OOD dataset.
Following MOS \citep{DBLP:conf/cvpr/HuangL21}, these OOD data are randomly selected from the categories that do not overlap with ImageNet-1k \citep{DBLP:conf/cvpr/DengDSLL009}.
Furthermore, some of the ablation experiments are conducted using ImageNet-100 as the ID data. This dataset follows the configuration of MCM \citep{DBLP:conf/nips/MingCGSL022} which selects 100 classes from ImageNet-1k as the ID data.

\noindent\textbf{Pre-trained Model.}
In our experiments,  we employ CLIP-B/16 as the pre-trained model for OOD prompt learning. 
Concretely, we choose CLIP-B/16, which consists of a ViT-B/16 Transformer as the image encoder and a self-attention Transformer as the text encoder. CLIP is one of the most popular pre-trained models, which learns from large-scale image-text datasets to create a shared embedding space where images and their associated text descriptions are represented coherently. By using contrastive learning, CLIP ensures similar image-text pairs closer together and dissimilar pairs farther apart, allowing it to understand the semantic relationships between visuals and language. In our experiment, we keep all the network parameters of CLIP fixed, including both the image encoder and the text encoder. We only update the embedding layer on the text input side, following the approach of prompts learning. 

\noindent\textbf{Implementation Details.}
For few-shot training, it is necessary to randomly select a certain number of samples from each class in the complete training data to form the training set. For example, we randomly choose one (one-shot) or four samples (four-shot) from each class in ImageNet-1k. When constructing ID and OOD data, we conduct $M$ (256 in our experiment) random crops on each sample, and choose the top $Q$ (32 in our experiment) and bottom $Q$ samples based on the similarity to the manually prompts. 
For ID prompts, there is only one learnable prompt per class, and class name information is retained. 
For OOD prompts, we set their total number to $C$ (100 in our experiment), and class name information is not retained. 
We set $\displaystyle \lambda_1$ to 0.3, $\displaystyle \lambda_2$ to 0.2, and use AdamW \citep{DBLP:conf/iclr/LoshchilovH19} as the optimizer. Other hyperparameters settings are as follows: training epoch\ =\ 3, learning rate\ =\ 0.005, batch size\ =\ 1, and token length $L$\ =\ 16.

\noindent\textbf{Competing Methods.} We compare our method to several OOD detection works, including fully supervised, zero-shot, and few-shot approaches. For fully supervised methods, we follow the same setting as NPOS \citep{DBLP:conf/iclr/TaoDZ023}, and compare with MSP \citep{DBLP:conf/iclr/HendrycksG17}, Fort/MSP \citep{DBLP:conf/nips/FortRL21}, Energy score \citep{DBLP:conf/nips/LiuWOL20}, ODIN score \citep{DBLP:conf/iclr/LiangLS18}, VOS \citep{DBLP:conf/iclr/DuWCL22}, and NPOS \citep{DBLP:conf/iclr/TaoDZ023}. 
For zero-shot methods, we select MCM \citep{DBLP:conf/nips/MingCGSL022} for comparison. 
For few-shot methods, we compare with CoOp \citep{DBLP:journals/ijcv/ZhouYLL22} and LoCoOp \citep{DBLP:journals/corr/abs-2306-01293}. For fairness, all methods are trained using the same pre-trained model (CLIP/ViT-B/16), and we reproduce some results from NPOS \citep{DBLP:conf/iclr/TaoDZ023} and LoCoOp \citep{DBLP:journals/corr/abs-2306-01293}.

\noindent\textbf{Evaluation Metrics.}
We adopt the following evaluation metrics that are commonly used in OOD detection: (1) the false positive rate of OOD examples when the true positive rate of in-distribution examples is at 95\% (FPR95); (2) the area under the receiver operating characteristic curve (AUROC); (3) ID classification accuracy (ID ACC).

\setlength{\abovecaptionskip}{0pt}
\setlength{\belowcaptionskip}{-3pt}
\begin{table*}[t]
\caption{Zero shot and few-shot OOD detection performance for ImageNet-1k \citep{DBLP:conf/cvpr/DengDSLL009} as ID. ViT-B/16 is an image encoder for CLIP-B/16, ViT-B$\displaystyle ^{+}$/16 uses the text encoder of CLIP-B/16 for initialization. 
}
\label{result_1k}
\renewcommand{\arraystretch}{1.0}
\begin{tabularx}{\textwidth}{
p{2.2cm}p{1.9cm}
c@{\hspace{4pt}}c
c@{\hspace{4pt}}c
c@{\hspace{4pt}}c
c@{\hspace{4pt}}c
}

\hline
~ & ~ 
& \multicolumn{8}{c}{OOD Dataset}\\
Method & Backbone
& \multicolumn{2}{c}{iNaturalist} 
& \multicolumn{2}{c}{Places} 
& \multicolumn{2}{c}{Texture} 
& \multicolumn{2}{c}{Average}\\
\cmidrule(r){3-4}  \cmidrule(r){5-6} \cmidrule(r){7-8} \cmidrule(r){9-10}


~ & ~
& \multicolumn{1}{c}{\fontsize{9}{14}\selectfont FPR95↓}
& \multicolumn{1}{c}{\fontsize{9}{14}\selectfont AUROC↑}
& \multicolumn{1}{c}{\fontsize{9}{14}\selectfont FPR95↓}
& \multicolumn{1}{c}{\fontsize{9}{14}\selectfont AUROC↑}
& \multicolumn{1}{c}{\fontsize{9}{14}\selectfont FPR95↓}
& \multicolumn{1}{c}{\fontsize{9}{14}\selectfont AUROC↑}
& \multicolumn{1}{c}{\fontsize{9}{14}\selectfont FPR95↓}
& \multicolumn{1}{c}{\fontsize{9}{14}\selectfont AUROC↑}\\

\hline
~ & ~ 
& \multicolumn{8}{c}{Full/Sub Data Fine-tune}\\
MSP \citep{DBLP:conf/iclr/HendrycksG17} &  CLIP-B/16 
            & 40.89 & 88.63 & 67.90 & 80.14 & 64.96 & 78.16 & 57.92 & 82.31 \\
Energy \citep{DBLP:conf/nips/LiuWOL20} &  CLIP-B/16
            & 29.75 & 94.68 & 56.40 & 85.60 & 51.35 & 88.00 & 45.83 & 89.43 \\
ODIN \citep{DBLP:conf/iclr/LiangLS18} &  CLIP-B/16
            & 30.22 & 94.65 & 55.06 & 85.54 & 51.67 & 87.85 & 45.65 & 89.35 \\
Fort/MSP \citep{DBLP:conf/nips/FortRL21} &  ViT-B/16
            & 54.05 & 87.43 & 72.98 & 78.03 & 68.85 & 79.06 & 65.29 & 81.51 \\
VOS \citep{DBLP:conf/iclr/DuWCL22} &  ViT-B/16
            & 31.65 & 94.53 & 41.62 & 90.23 & 56.67 & 86.74 & 43.31 & 90.50 \\
NPOS \citep{DBLP:conf/iclr/TaoDZ023} &  ViT-B$\displaystyle ^{+}$/16
            & 16.58 & 96.19 & 45.27 & 89.44 & 46.12 & 88.80 & 35.99 & 91.48 \\
\hline
~ & ~ 
& \multicolumn{8}{c}{Zero-shot}\\
MCM \citep{DBLP:conf/nips/MingCGSL022} &  CLIP-B/16
            & 30.91 & 94.61 & 44.69 & 89.77 & 57.77 & 86.11 & 44.46 & 90.16 \\
\hline
~ & ~ 
& \multicolumn{8}{c}{One-shot}\\
CoOp \citep{DBLP:journals/ijcv/ZhouYLL22} &  CLIP-B/16
            & 43.38 & 91.26 & 46.68 & 89.09 & 50.64 & 87.83 & 46.90 & 89.39 \\
LoCoOp \citep{DBLP:journals/corr/abs-2306-01293} &  CLIP-B/16
            & 38.49 & 92.49 & \textbf{39.23} & 91.07 & 49.25 & 89.13 & 42.32 & 90.90 \\
\rowcolor{gray!20}
Ours &  CLIP-B/16
            & \textbf{14.57} & \textbf{97.35} & 41.74 & \textbf{91.15} & \textbf{26.77} & \textbf{94.38} & \textbf{27.69} & \textbf{94.29} \\
\hline
~ & ~ 
& \multicolumn{8}{c}{Four-shot}\\
CoOp \citep{DBLP:journals/ijcv/ZhouYLL22} &  CLIP-B/16
            & 35.36 & 92.60 & 45.38 & 89.15 & 43.74 & 89.68 & 41.49 & 90.48 \\
LoCoOp \citep{DBLP:journals/corr/abs-2306-01293} &  CLIP-B/16
            & 29.45 & 93.93 & \textbf{41.13} & 90.32 & 44.15 & 90.54 & 38.24 & 91.60 \\
\rowcolor{gray!20}
Ours &  CLIP-B/16
            & \textbf{  8.98} & \textbf{98.19} & 44.00 & \textbf{90.57} & \textbf{25.27} & \textbf{94.32} & \textbf{26.08} & \textbf{94.36} \\
\hline
\end{tabularx}
\end{table*}

\subsection{Results}

\ 

\begin{figure}[tp]
    \vspace{0.2em}
  \centering
  \includegraphics[width=0.43\textwidth]{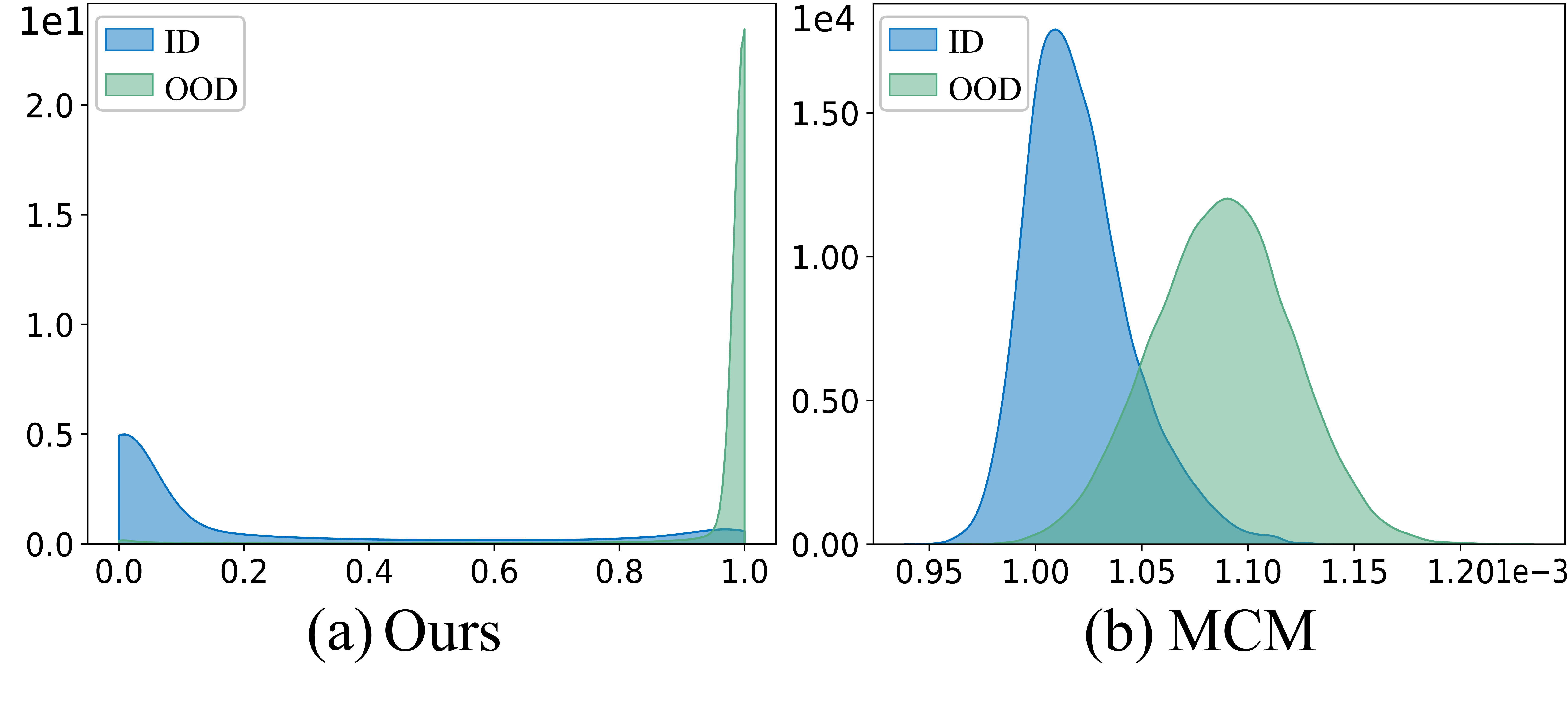}
  \caption{
  Density of the obtained ID and OOD score with the proposed method (left) and MCM \cite{DBLP:conf/nips/MingCGSL022} (right).}
    \label{fig_distribution}
\end{figure}

Table \ref{result_1k} shows our main comparison results, which demonstrate that using our method can achieve better OOD detection performance, outperforming most comparisons. More importantly, our method still has good results in 1-shot setting even compared to those methods that require full data. Specifically, in the 4-shot setting, we obtain 26.08\% in terms of FPR95 and 94.36\% in terms of AUROC on average, implying a reduction of 12.16\% and an improvement of 2.76\%, respectively compared to the best-performing method under the same settings. 
Fig.~\ref{fig_distribution} shows a comparison between our method and MCM on the iNaturalist dataset. Our approach demonstrates superior performance with a significantly larger discrepancy between ID and OOD. This suggests that MCM is more sensitive to threshold when distinguishing between ID and OOD, whereas our method allows a more intuitive distinction between ID and OOD. Furthermore, as shown in Table~\ref{result_acc}, our method outperforms other few-shot methods, achieving superior classification results on ID data at 68.28\%.

\begin{table}[t]
\vspace{0.2em}
\caption{ID accuracy on ImageNet-1k \citep{DBLP:conf/cvpr/DengDSLL009}.}
\label{result_acc}
\renewcommand{\arraystretch}{1.0}
\begin{tabularx}{0.47\textwidth}{
p{1.9cm}
cccc
}
\hline

Method
& \multicolumn{1}{c}{\fontsize{9}{14}\selectfont ID acc}
& \multicolumn{1}{c}{\fontsize{9}{14}\selectfont Full Data} 
& \multicolumn{1}{c}{\fontsize{9}{14}\selectfont Zero-shot} 
& \multicolumn{1}{c}{\fontsize{9}{14}\selectfont One-shot} \\
\hline
VOS \citep{DBLP:conf/iclr/DuWCL22} & 79.64 & \checkmark &   &   \\
NPOS \citep{DBLP:conf/iclr/TaoDZ023} & 79.42 & \checkmark &   &   \\
\hline
MCM \citep{DBLP:conf/nips/MingCGSL022} & 67.01 &   & \checkmark &   \\
CoOp \citep{DBLP:journals/ijcv/ZhouYLL22} & 66.23 &   &   & \checkmark \\
LoCoOp \citep{DBLP:journals/corr/abs-2306-01293} & 66.88 &   &   & \checkmark \\
Ours & 68.28 &   &   & \checkmark \\
\hline
\end{tabularx}
\end{table}

\setlength{\belowcaptionskip}{-15pt}
\begin{table}[t]
\vspace{1.3em}
\caption{OOD detection performance  for ImageNet-1k as ID, SUN \citep{DBLP:conf/cvpr/XiaoHEOT10} as OOD.}
\label{result_sun}
\renewcommand{\arraystretch}{1.0}
\begin{tabularx}{0.47\textwidth}{
p{2.05cm}
c@{\hspace{4pt}}c
c@{\hspace{4pt}}c
}

\hline
~ & \multicolumn{4}{c}{SUN}\\

Method 
& \multicolumn{2}{c}{One-shot} 
& \multicolumn{2}{c}{Four-shot}\\

\cmidrule(r){2-3}  \cmidrule(r){4-5}

~ 
& \multicolumn{1}{c}{\fontsize{8}{14}\selectfont FPR95↓}
& \multicolumn{1}{c}{\fontsize{8}{14}\selectfont AUROC↑}
& \multicolumn{1}{c}{\fontsize{8}{14}\selectfont FPR95↓}
& \multicolumn{1}{c}{\fontsize{8}{14}\selectfont AUROC↑}\\

\hline

CoOp \citep{DBLP:journals/ijcv/ZhouYLL22}
& 38.53 & 91.95 & 37.06 & 92.27 \\
LoCoOp \citep{DBLP:journals/corr/abs-2306-01293}
& 33.27 & 93.67 & 33.06 & 93.24 \\
Ours
& 44.02 & 91.08 & 42.03 & 91.64 \\
\hline
\end{tabularx}
\vspace{-0.5em}
\end{table}

\begin{figure}[tp]
  \centering
  \includegraphics[width=0.42\textwidth]{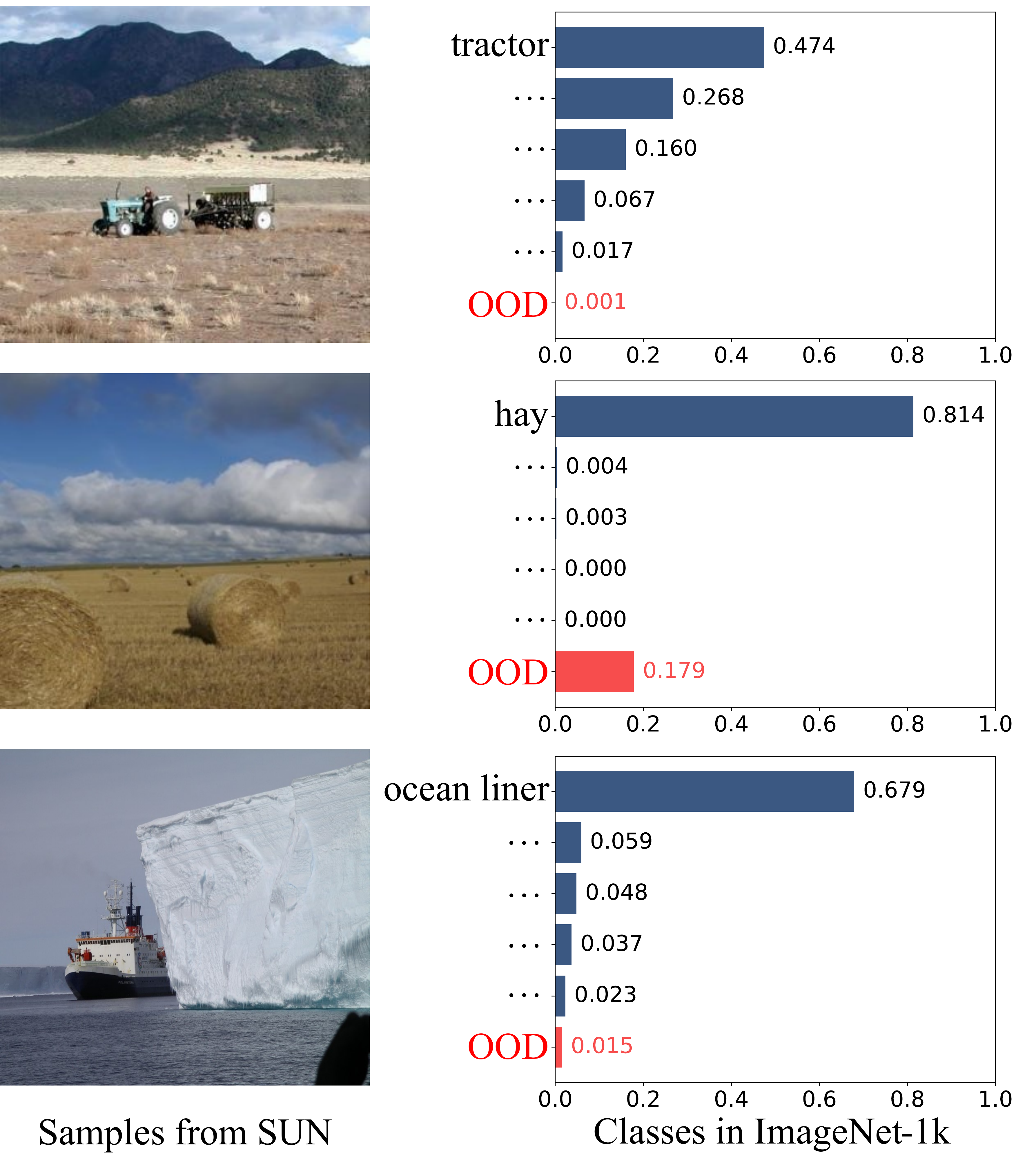}
  \caption{Samples from the SUN dataset that may be semantically identical to classes in ImageNet-1K, yet these samples are still considered as OOD during evaluation, which may result in a performance reduction.}
    \label{fig_SUN_samples}
\end{figure}

\noindent\textbf{Discussion on the SUN dataset.} 
We also conduct an evaluation on the SUN dataset~\citep{DBLP:conf/cvpr/XiaoHEOT10} as OOD data, and the results are shown in Table \ref{result_sun}. The results indicate that our method performs not well on the SUN dataset. To investigate the reasons, we conduct a detailed examination of the SUN dataset. Afterward, we find that some samples belong actually ID classes (as shown in Fig.~\ref{fig_SUN_samples}), but they are labeled as OOD. 
To investigate whether the observed case is a prevalent phenomenon in the SUN dataset, we conduct a more detailed analysis. We randomly select 400 samples from the SUN dataset and observe whether they belong to the ID category. We find that among these samples, 145 belong to the ID category. 
This investigation implies the following fact: the SUN dataset may require more detailed annotation and filtering to be suitable as OOD data for testing the OOD detection performance (Places \citep{DBLP:journals/pami/ZhouLKO018} might also have similar issues, but due to the space limitation, we leave this in future).

\setlength{\belowcaptionskip}{-12pt}

\setlength{\belowcaptionskip}{-3pt}
\begin{table*}[t]
\caption{Ablation study by constructing different outlier training data. The experimental results show that the proposed method of constructing outlier has achieved significant improvements.}
\label{result_data}
\renewcommand{\arraystretch}{1.0}
\begin{tabularx}{\textwidth}{
p{2.65cm}
c@{\hspace{4pt}}c
c@{\hspace{4pt}}c
c@{\hspace{4pt}}c
c@{\hspace{4pt}}c
c@{\hspace{4pt}}c
}
\hline
~ & \multicolumn{10}{c}{OOD Dataset}\\
Outlier (train)
& \multicolumn{2}{c}{iNaturalist} 
& \multicolumn{2}{c}{Places} 
& \multicolumn{2}{c}{Texture} 
& \multicolumn{2}{c}{SUN} 
& \multicolumn{2}{c}{Average}\\
\cmidrule(r){2-3}  \cmidrule(r){4-5} \cmidrule(r){6-7} \cmidrule(r){8-9} \cmidrule(r){10-11}

~ 
& \multicolumn{1}{c}{\fontsize{8}{14}\selectfont FPR95↓}
& \multicolumn{1}{c}{\fontsize{8}{14}\selectfont AUROC↑}
& \multicolumn{1}{c}{\fontsize{8}{14}\selectfont FPR95↓}
& \multicolumn{1}{c}{\fontsize{8}{14}\selectfont AUROC↑}
& \multicolumn{1}{c}{\fontsize{8}{14}\selectfont FPR95↓}
& \multicolumn{1}{c}{\fontsize{8}{14}\selectfont AUROC↑}
& \multicolumn{1}{c}{\fontsize{8}{14}\selectfont FPR95↓}
& \multicolumn{1}{c}{\fontsize{8}{14}\selectfont AUROC↑}
& \multicolumn{1}{c}{\fontsize{8}{14}\selectfont FPR95↓}
& \multicolumn{1}{c}{\fontsize{8}{14}\selectfont AUROC↑}\\
\hline
~ & \multicolumn{10}{c}{Related outlier}\\
iNaturalist \citep{DBLP:conf/cvpr/HornASCSSAPB18}
& \textcolor{gray}{  1.57} & \textcolor{gray}{99.62}	& 41.57 & 90.11	& 81.29 & 69.62 & 40.05 & 89.26	& 41.12 & 87.15\\
Places \citep{DBLP:journals/pami/ZhouLKO018}
& 30.68 & 95.17	& \textcolor{gray}{10.64} & \textcolor{gray}{97.91}	& 75.57 & 77.19	& \textbf{22.62} & \textbf{95.14}	& 34.88 & 91.35\\
Texture \citep{DBLP:conf/cvpr/CimpoiMKMV14}
& 15.24 & 97.27	& 22.07 & 95.60	& \textcolor{gray}{33.51} & \textcolor{gray}{93.03}	& 41.45 & 90.40 & 28.07 & 94.07\\
SUN \citep{DBLP:conf/cvpr/XiaoHEOT10}
& 31.14 & 94.72	& 24.04 & 94.70	& 88.21 & 71.02	& \textcolor{gray}{11.06} & \textcolor{gray}{97.65}	& 38.61 & 89.52\\
\hline
~ & \multicolumn{10}{c}{Unrelated outlier}\\
CUB \citep{WahCUB_200_2011}
& 60.88 & 89.63	& 59.40 & 87.34	& 48.63 & 89.00	& 78.39 & 78.62	& 61.83 & 86.15\\
Gaussian Noise 
& 52.18 & 91.11	& 80.53 & 75.13	& 33.95 & 91.40	& 61.94 & 85.09	& 57.15 & 85.68\\
\hline
\rowcolor{gray!20}
~ & \multicolumn{10}{c}{Our outlier}\\
\rowcolor{gray!20}
One-shot    
& 10.82 & 97.84 & 26.60 & 95.07 & 20.41 & 96.16 & 41.94 & 91.43 & 24.94 & 95.12\\
\rowcolor{gray!20}
Four-shot    
& \textbf{  1.62} & \textbf{99.60} & \textbf{20.81} & \textbf{96.05} & \textbf{13.55} & \textbf{97.30} & 29.53 & 93.70 & \textbf{16.38} & \textbf{96.66}\\
\hline
\end{tabularx}
\end{table*}

\noindent\textbf{Discussion of performance differences.}
We briefly analyze the performance differences of our method across different OOD datasets.
For example, our method exhibits significant performance improvement on iNaturalist \citep{DBLP:conf/cvpr/HornASCSSAPB18} and Texture \citep{DBLP:conf/cvpr/CimpoiMKMV14} datasets. 
The possible reason is that the cropped samples are more likely to contain image textures, plants, and animals in the background, making them correlated with ID classes. iNaturalist consists of various types of plants and animals, and Texture consists of natural textures. Therefore, our approach exhibits a greater improvement on these two datasets. In contrast, SUN \citep{DBLP:conf/cvpr/XiaoHEOT10} and Places \citep{DBLP:journals/pami/ZhouLKO018} primarily consist of scene-based data, typically lacking specific objects (e.g., containing multiple objects). Therefore, our approach shows limited performance improvement on these two datasets.

\subsection{Ablation Study}

\setlength{\belowcaptionskip}{-10pt}
\begin{table}[t]
\vspace{1.3em}
\caption{Ablation study of the effectiveness of \idlike prompts and the proposed OOD score function.}
\label{result_OOD_prompts}
\renewcommand{\arraystretch}{1.0}
\begin{tabularx}{0.47\textwidth}{
p{2.05cm}
c@{\hspace{4pt}}c
c@{\hspace{4pt}}c
}

\hline
~ & \multicolumn{4}{c}{Average}\\

Method 
& \multicolumn{2}{c}{One-shot} 
& \multicolumn{2}{c}{Four-shot}\\

\cmidrule(r){2-3}  \cmidrule(r){4-5}

~ 
& \multicolumn{1}{c}{\fontsize{8}{14}\selectfont FPR95↓}
& \multicolumn{1}{c}{\fontsize{8}{14}\selectfont AUROC↑}
& \multicolumn{1}{c}{\fontsize{8}{14}\selectfont FPR95↓}
& \multicolumn{1}{c}{\fontsize{8}{14}\selectfont AUROC↑}\\

\hline

ID prompts
& 49.57 & 88.25 & 49.28 & 87.86 \\
All prompts
& 44.18 & 89.80 & 42.51 & 89.81 \\
Ours
& 27.69 & 94.29 & 26.08 & 94.36 \\
\hline
\end{tabularx}
\vspace{-0.2em}
\end{table}

\noindent\textbf{The effectiveness of \idlike prompts.} 
We compare the average performance of our method in OOD detection on ImageNet-1k under different conditions. We show the MCM scores when only ID prompts are presented, MCM scores when both ID prompts and OOD prompts are used simultaneously, and the proposed score when all prompts are provided. As shown in Table \ref{result_OOD_prompts}, we observe that with the introduction of OOD prompts, the OOD detection performance significantly improves. This indicates that OOD prompts can promisingly distinguish between ID and OOD samples.

\noindent\textbf{The effectiveness of our outliers. }
To show the effectiveness of the outliers constructed, we conduct the following experiments. Specifically, we using different additional outliers in training to investigate the improvement of our constructed outliers. Furthermore, we categorize the auxiliary outliers into ``Related outlier'', ``Unrelated outlier'' and ``Our outlier''. Concretely, ``Related outlier'' are selected from the challenging OOD samples (those with high MSP scores \citep{DBLP:conf/iclr/HendrycksG17}). "Unrelated outlier" are selected from OOD datasets that are unrelated to the ID data. The results are shown in Table \ref{result_data}. Partial experimental results are in gray because it is unfair to compare them, since the outliers used during training and the OOD in testing come from the same distribution.  Firstly, it is observed that models trained with our generated outliers outperform those trained with other outliers. This indicates the effectiveness of the outliers generated by our method. Secondly, we observe that the performance of models trained with ``Related outlier'' is generally better than those trained with ``Unrelated outlier''. This supports that outliers related to the ID can indeed help the model in learning a better decision boundary between the ID and OOD. The overall result strongly validates our \idlike outliers are quite effective and reasonable.

\begin{figure}[tp]
    \begin{center}
    \includegraphics[width=0.4\textwidth, height=0.2\textwidth]{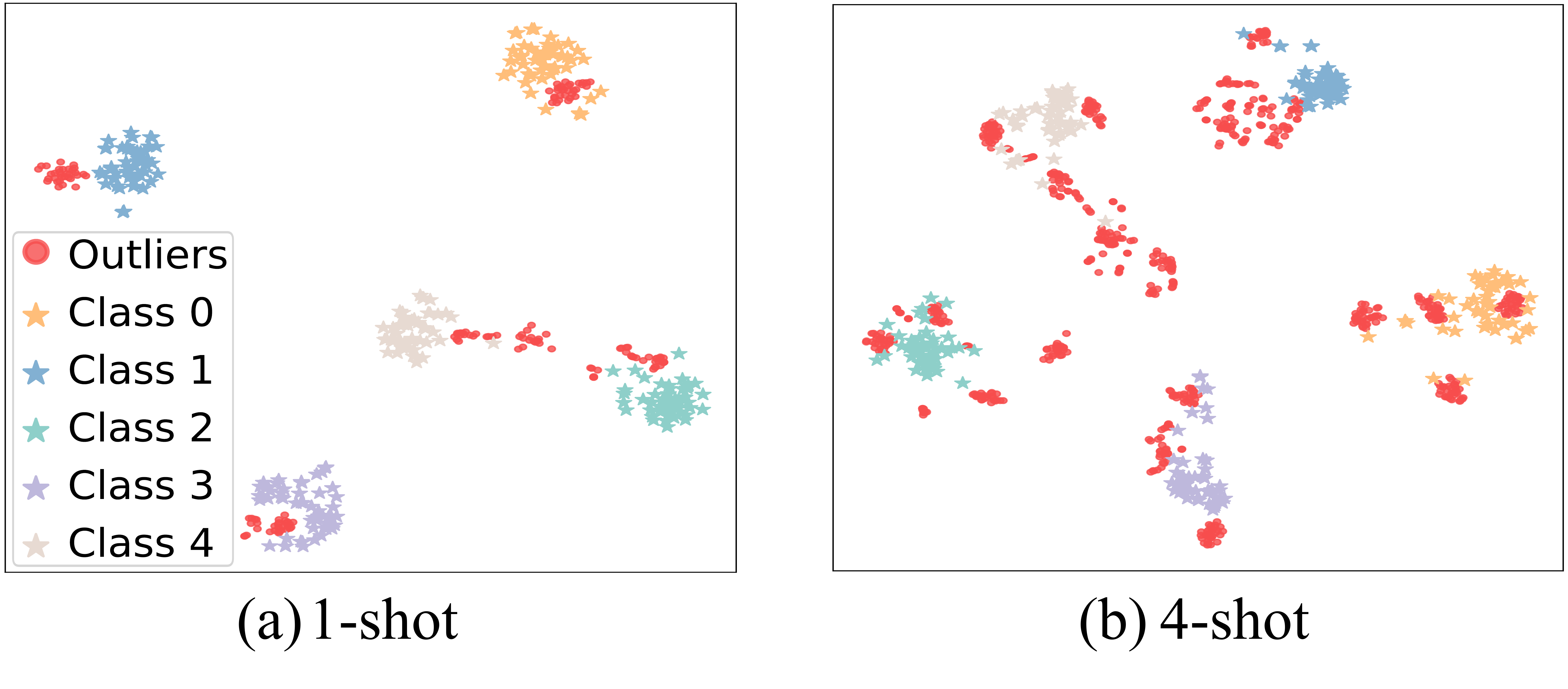}
    \end{center}
    \caption{
    The obtained representations visualization of \idlike OOD samples and ID samples under 1-shot and 4-shot settings. The representation of the obtained \idlike OOD sample is close to the ID sample. 
    }
    \label{fig_diversity_2}
    \vspace{-0.5em}
\end{figure}

Furthermore, we utilize t-SNE \citep{van2008visualizing} for visualization to illustrate the correlation between the outliers generated by our method and the ID samples. We employ a number of ID samples along with the outliers constructed based on the few-shot setting for visualization. For example, under the 1-shot setting, we use only one sample form each ID class to construct outliers, while the ID samples consist of a large number of samples form different ID classes. As shown in Fig.~\ref{fig_diversity_2}, the results show that even the generated outliers are from a quite small number of ID samples, they can also be correlated with the majority of ID samples. Moreover, with the increase of ID samples used in constructing outliers, both the number and the diversity of the \idlike outliers also increase.

\begin{figure}[tp]
    \begin{center}
    \includegraphics[width=0.47\textwidth]{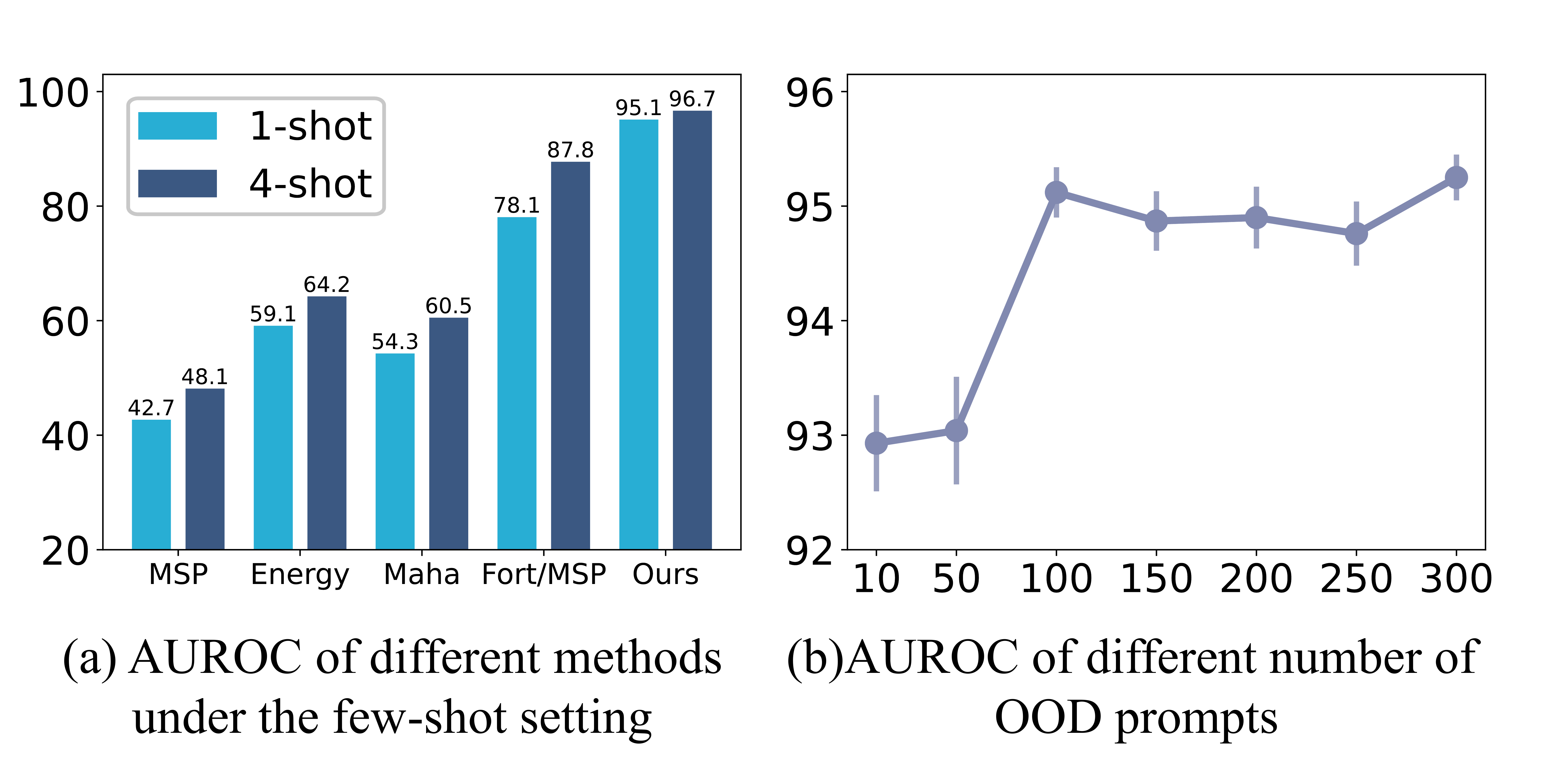}
    \end{center}
    \caption{
    Left: Performance in terms of  AUROC of different methods trained on our constructed \idlike OOD dataset.
    Right: Performance in terms of  AUROC at different number of OOD prompts during training.
    }
    \label{fig_methods_K}
\end{figure}

\noindent\textbf{The effectiveness of prompt learning. }
To show the advantages of prompt learning under few-shot setting, we train various models with our generated outliers (ImageNet-100 as ID), including fine-tuning the full model \citep{DBLP:conf/iclr/HendrycksG17, DBLP:conf/nips/LiuWOL20}, fine-tuning the last layer \citep{DBLP:conf/nips/FortRL21}, training free \citep{DBLP:conf/nips/LeeLLS18}, and prompt learning (ours). 
As shown in Fig.~\ref{fig_methods_K} (a), the results show that fine-tuning the full model performs worse in both 1-shot and 4-shot settings. The main reason is that these methods typically require abundant data for fine-tuning the model. Moreover, the better performance of all methods under 4-shot over 1-shot settings also validates this. 
Fort/MSP \citep{DBLP:conf/nips/FortRL21}, which fine-tunes only the last layer of the model, performs better than fine-tuning the full model. This is because it preserves the majority of the model's prior knowledge, thereby reducing the dependence on the quantity of training data. However, it only utilizes the image encoder and does not leverage the pre-trained model's prior knowledge in text, thereby limiting the performance of CLIP. 
\citet{DBLP:conf/nips/LeeLLS18} does not fine-tune the model, but it only utilizes the image encoder without fully leveraging CLIP's prior knowledge, limiting its performance. 
In contrast, prompt learning often utilizes both the image encoder and text encoder, leveraging the full prior knowledge of the pre-trained model. Therefore, for the limited amount of ID data, the performance is significant superior compared to existing methods. 

\setlength{\abovecaptionskip}{-10pt}
\setlength{\belowcaptionskip}{-6pt}

\setlength{\abovecaptionskip}{-5pt}
\setlength{\belowcaptionskip}{-10pt}

\setlength{\abovecaptionskip}{5pt}
\setlength{\belowcaptionskip}{-6pt}

\noindent\textbf{The effectiveness of different quantities of \idlike prompts (OOD prompts). }
We test the impact of different quantities of OOD prompts for OOD detection performance. We set different values for $C$ (10, 50, 100, 150, 200, 250, 300) and train the model using ImageNet-100 as ID data. The results are shown in Fig.~\ref{fig_methods_K} (b). The results demonstrate that as the number of OOD prompts increases, the OOD detection performance is also  improved and tends to be stable. The underlying reason is that the expressive capacity is highly related to the number of prompts. Therefore, when ID data is complex, more prompts are required to characterize OOD samples related to the ID samples.

\section{Conclusion}
\label{sec:conclusion}

\ 

In this work, we propose a novel few-shot prompt learning method for out-of-distribution detection using pre-trained visual-language models. Our method introduces ID-like prompts and constructs outliers highly correlated with ID data from the training samples. By aligning ID-like prompts with the constructed outliers, we explore ID-like regions within the text feature space that are highly correlated with ID but do not belong to the ID. Our method elegantly addresses the key limitations in previous OOD detection methods, i.e., the challenge of constructing challenging outliers without auxiliary outliers and with a limited number of ID samples. Additionally, the introduction of the ID-like prompt provides a more effective way for the model to identify challenging OOD data. In challenging real-world OOD detection tasks, our method outperforms existing approaches. We conducted various ablation experiments to demonstrate the effectiveness of our approach. We hope that our work could inspire more future research on few-shot OOD detection based on prompts learning.


\clearpage

{\small
\bibliographystyle{ieeenat_fullname}
\bibliography{11_references}
}


\end{document}


\title{\paperTitle}
\author{\authorBlock}
\maketitlesupplementary

\section{Appendix Section}

\subsection{Effectiveness of OOD prompts in OOD detection task. }

\ 

Given we obtain the similarity of the test sample to all prompts $\displaystyle s=sim(z^{img}, z^{txt})$. 
Assume all prompts have been well optimized. Assuming there are two test samples, one being ID and one being highly ID-related OOD, 
$\displaystyle s_{in,in}=sim(z^{img}_{in}, z^{txt}_{in})$, 
$\displaystyle s_{out,in}=sim(z^{img}_{out}, z^{txt}_{in})$, 
$\displaystyle s_{in,out}=sim(z^{img}_{in}, z^{txt}_{out})$, 
$\displaystyle s_{out,out}=sim(z^{img}_{out}, z^{txt}_{out})$. 
Let $\displaystyle s^{max}$ be the highest similarity value, $\displaystyle s^{other}$ be the values other than the highest. 
We have the following cases: 
$\displaystyle s^{max}_{in,in}=s^{max}_{out,in}$, 
$\displaystyle s^{other}_{in,in} \approx s^{other}_{out,in}$, 
$\displaystyle s_{in,out}<s_{out,out}$. 

\ 

When using MCM score: 
\begin{itemize}[leftmargin=15pt]
    \item 
    For ID sample:
    
    $\displaystyle
        S^{mcm}_{in}=\frac
            {
                exp(s^{max}_{in,in})
                }
            {
                exp(s^{max}_{in,in}) + 
                \textcolor{blue!70}{\Sigma{exp(s^{other}_{in,in})}}
            }$.
    \item 
    For highly ID-related OOD sample:
    
    $\displaystyle
        S^{mcm}_{out}=\frac
            {
                exp(s^{max}_{out,in})
                }
            {
                exp(s^{max}_{out,in}) + 
                \textcolor{blue!70}{\Sigma{exp(s^{other}_{out,in})}}
            }$.
\end{itemize}

$\because \displaystyle s^{max}_{in,in}=s^{max}_{out,in}$, 
$\displaystyle \textcolor{blue!70}{s^{other}_{in,in}} \approx \textcolor{blue!70}{s^{other}_{out,in}}$, 

$\therefore \displaystyle S^{mcm}_{in} \approx S^{mcm}_{out}$. 

\ 

When using our score: 
\begin{itemize}[leftmargin=15pt]
    \item 
    For ID sample:
    
    $\displaystyle
        S^{mcm}_{in}=\frac
        {
            exp(s^{max}_{in,in}) + 
            \textcolor{blue!70}{\Sigma{exp(s^{other}_{in,in})}}
            }
        {
            exp(s^{max}_{in,in}) + 
            \textcolor{blue!70}{\Sigma{exp(s^{other}_{in,in})}} + 
            \textcolor{red!70}{\Sigma{exp(s_{in,out})}}
        }$.
    \item 
    For highly ID-related OOD sample:
    
    $\displaystyle
        S^{mcm}_{out}=\frac
        {
            exp(s^{max}_{out,in}) + 
            \textcolor{blue!70}{\Sigma{exp(s^{other}_{out,in})}}
            }
        {
            exp(s^{max}_{out,in}) + 
            \textcolor{blue!70}{\Sigma{exp(s^{other}_{out,in})}} + 
            \textcolor{red!70}{\Sigma{exp(s_{out,out})}}
        }$.
\end{itemize}

$\because \displaystyle s^{max}_{in,in}=s^{max}_{out,in}$, 
$\displaystyle \textcolor{blue!70}{s^{other}_{in,in}} \approx \textcolor{blue!70}{s^{other}_{out,in}}$, 
$\displaystyle \textcolor{red!70}{s_{in,out}}<\textcolor{red!70}{s_{out,out}}$, 

$\therefore \displaystyle S^{ours}_{in} > S^{ours}_{out}$. 

Therefore, in detecting OOD that is highly similar to ID, our method is more effective than MCM.

\clearpage

{\small
\bibliographystyle{ieee_fullname}
\bibliography{11_references}
}